# Why did the accident happen? A norm-based reasoning approach

Farid NOUIOUA
Laboratoire d'Informatique de Paris-Nord  UMR 7030 du CNRS
Avenue Jean-Baptiste Clément, F-93430 Villetaneuse, France
e-mail : nouiouaf@lipn.univ-paris13.fr

## I. Introduction

To understand well a text written in natural language (*NL*), we need our knowledge about the *norms* of its domain. By the word "*norm*", we mean here the normal and expected course of events in the absence of exceptions [6]. This type of knowledge enables us to infer richer conclusions than those given by means of truth-preserving entailments, for example, from the text: **"Mon véhicule se trouvait arrêté à un stop, quand un véhicule m'a heurté à l'arrière"**, (**My vehicle was stopped at a stop sign, when a vehicle struck me at the back**). Norms provide conclusions like: *vehicle A and me were in the same file and direction*, *vehicle A had to stop to avoid the shock…* None of these conclusions is explicit. However, any reader infers them immediately. Conclusions obtained by using norms can in general be defeasible, but they are accepted as long as the text does not contradict them. Often, narrative texts do not describe norms explicitly. They focus rather on their violations, by describing generally abnormal situations. In the light of this main remark, our goal consists in looking for the *cause* of an accident from its textual description by hypothesizing that the searched cause (called the *primary anomaly*) is the violation of the most specific norm in the text [3]. The other violations of norms result from the first one and are called *derived anomalies*. We are working on a corpus of 60 car crash reports written in French. Each report is a small text describing briefly the circumstances of an accident. To validate our approach, the reasoning system must find for each text the same answer given by an ordinary human reader to the question: *"what is the most specific violated norm which can considered as the plausible cause of the accident ?"*. These answers are determined manually for each text at the beginning of the process.

## II. Overall architecture

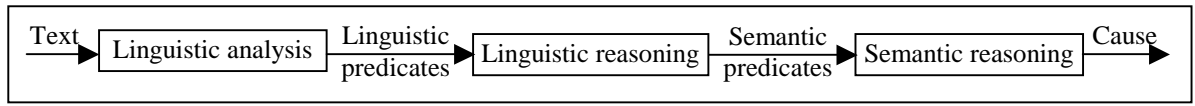

As shown in the figure above, several steps are required in the process of finding the cause. We will explain the role of each step further. We just notice here, that in our methodology, we have started by developing the *semantic reasoning* before dealing with the *linguistic* one. This enabled us to determine a reasonable set of *semantic predicates* (around 50) in terms of which the *linguistic reasoning* should express what is needed, and only what is needed from the explicit content of the text. This methodology enables the reasoning process to deal only with relevant linguistic phenomena. In this work, we focus on the extraction a set of syntactical relations between the words of the text and then we use a reasoning process to transform these relations into a set of semantic predicates.

## III. Linguistic analysis

The *tree tagger*[1] is applied to the text. The result is, then, passed to a parser which uses a *context free grammar* enhanced with appropriate *semantic actions* to produce a set of linguistic predicates. These predicates reflect syntactic relations between relevant words of the text.  At the end of this step, we obtain from our example:

*qualif_n(véhicule, Mon), subject(se_trouver, véhicule), qualif(trouver, arrêté), compl_v(à, trouver, stop), compl_v(quand, trouver, heurter), subject(heurter, véhicule), object(heurter, m'), compl_v(à, heurter, arrière).*

## IV. Linguistic reasoning

The aim of the linguistic reasoning is to transform the linguistic predicates into semantic ones which express the explicit content of the text. The main idea (The development of this step is still in progress) is to design general transformation rules based on a lexical semantic study of the words. Of course, rules of this kind are, in general, defeasible and one must handle their exceptions. That is why a non-monotonic approach is required at this level.

---

[1] http://www.ims.uni-stuttgart.de/projekte/corplex/TreeTagger/DecisionTreeTagger.html

The linguistic predicates obtained for the example in the previous step are transformed by the linguistic reasoning into the following semantic predicates (see the representation details in the following section)

*Holds(stop, A, 1) : the a*gent *A* is stopped at time *1*.

*Holds(stop_sign, A, 1) :* there is a stop sign for the agent *A* at time *1*.

*Holds(combine(bump, A), B,2) :* the agent *B* bumps the agent *A* at time *2*.

*Holds(combine(shock_pos, back), A, 2) :* the position of the shock of the agent *A* is its back.

## V. Semantic reasoning

The semantic predicates obtained are the input of the semantic reasoning step. This step uses inference rules based on our knowledge about norms of the road domain to enrich the initial conclusions by further implicit ones, and enables to detect the primary anomaly, which we consider as the cause of the accident. So, our common knowledge about the norms of the road domain are expressed by means of inference rules.

### V.1. Language

Before showing what our inference rules look like, let us give briefly the main ingredients of the logical representation language used (see [2] for more details).

Although we need some features that are normally treated by higher order logics, we have chosen, for efficiency reasons, to stay in a first order logic (*FOL*) framework. To do this, we use the usual reification technique to represent modalities and to quantify over predicate names. Thus, a binary predicate *P(X, Y)* is written *Holds(P, X, Y)*.

Temporal aspect is a central issue in causal reasoning [4]. To deal with this question, our approach is to decompose the scene of the accident into a succession of intervals characterized by the truth values of a set of literals. We add a parameter to each time-dependent predicate. This parameter represents the order number of the interval in which the corresponding predicate or its negation holds. Strictly speaking, the exact meaning of the temporal parameter T depends on the considered property: For properties such that "move", "stop", "control", … the parameter T represents the whole time interval T. Indeed, this type of properties are generally persistent i.e. they hold all along throughout the time interval T. For properties such that "starts", "bump", …, T represents rather a particular time point that belongs to the time interval T. To simplify, we will use the expression "at time T" with the two types of properties. Thus, the literal *Holds(P, A, T)* is true iff property *P* holds for agent *A* at time *T*. For predicates with more than two arguments, we use the binary function *combine* : the ternary *P(A, B, t)* is written *Holds(combine(P, A), B, t). combine(P, A)* is a composed property. To decide which argument will be in the function combine and which one stays in the predicates Holds, the criterion is that the second argument of Holds is the principal agent of the property whereas the other one is used to construct with the initial simple property a composed one. For exemple in *"A follows B at time T"*, the principal agent of the property "to *follow"* is *A,*  using the simple property *"follows"* and the argument *B*, we define the composed property *"following B"* expressed by: *combine(follows, B)*. The resulting predicate is then: *Holds(combine(follows, B), A, T)*.

In addition to the *Holds* predicate which expresses truth values, we need two modalities: the *Must* modality which expresses duties of agents and the *Able* modality which expresses their capacities: *Must(P, A, T)* (resp. *Able(P, A, T))* holds iff at time *T*, agent *A* has the duty (resp. is able) to reach the property *P*.

According to the previous representation, we define two forms for a primary anomaly:

$$Must(P, A, T) \land Able(P, A, T) \land Holds(P', A, T) \land Incompatible(P, P') \rightarrow P \_ Anomaly$$

$$Holds(combine(Disruptive \_ Factor, X), A, T) \rightarrow P \_ Anomaly$$

The first form expresses the fact that if at time *T*, the agent *A* has the duty to reach a property *P* and that it is able at this time to reach it, but that at time *T+1* a property *P'* incompatible with *P* holds, than there is a primary anomaly.

The second form of a primary anomaly is used to detect situations in which there is some disruptive factor that causes the accident and which generally can not be avoided by the agents. It is the case for example of the existence in the road of unforeseeable gravels or oil that cause loss of control to vehicles.

A derived anomaly differs from the first form of a primary anomaly only on the agent's ability:

$$Must(P, A, T) \land \neg Able(P, A, T) \land Holds(P', A, T) \land Incompatible(P, P') \rightarrow D \_ Anomaly$$

### V.2. Inference rules

Because norm-based conclusions are defeasible, a non-monotonic approach is required in writing the inference rules. We use Reiter's default logic [5]. The inference rules belong to two categories:

- Material implications of the form : $A \rightarrow B$, where $A$ is a conjunction of literals and B is a literal.

- Defaults; we have normal defaults of the form $\frac{A:B}{B}$ (abbreviated by writing *A : B)*, and semi-normal defaults of the form $\frac{A:B \wedge C}{B}$ (abbreviated by writing *A : B [C])*, where *A* and C are conjunctions of literals and B is a literal.

We define a kernel of a few semantic predicates such that all anomalies can be expressed in terms of these predicates. Thus, the reasoning process converges into the kernel predicates, and stops when the primary anomaly is found. The kernel contains six (reified) predicates:

*Holds(stop,A,T)* : the vehicle *A* is stopped at time *T*.

*Holds(run_slowly_enough,A,T)* : the speed of the vehicle *A* is adapted at time *T*.

*Holds(control, A, T)* : the vehicle *A* is controlled by its driver at time T.

*Holds(move_back, A, T)* : the vehicle *A* moves back at time *T*.

*Holds(combine(Disruptive_Factor, X), A, T)* : *there* is some disruptive factor *X* for the vehicle *A* at time *T*.

Let us now give some examples of inference rules and their application to our example to infer the primary anomaly. The semantic predicates obtained are those given in section IV.

The rule: $Holds(combine(bump,V),W,T) \rightarrow \neg Holds(stop,W,T)$ means that if W bumps V at time T, then W is not stopped at time T. Its application on the example gives : ¬*Holds(stop,B,2)* (*V = A, W = B, T = 2*).

The rule: $Holds(combine(bump,V),W,T) \rightarrow Holds(combine(shock,V),W,T)$ which means that if *W* bumps *V* at time *T* then there is a shock between *V* and *W* at this time *T* enable to deduce *Holds(combine(shock, A), B, 2)* (*V = A, W = B, T = 2*).

The following default expresses that in general, if there is a shock between *V* and *W* at time *T* and the shock position of *V* is its back, then *W* was the follower of *V* in the same file at time *T-1*. This rule is inhibited if *W* has not the control. By applying this default we infer : *Holds(combine(follows, A), B, 1)* (*V = A, W = B, T = 2*).

$Holds(combine(shock,V),W,T) \wedge Holds(combine(shock\_pos,back),V,T) :$
$$Holds(combine(follows,V),W,T-1)[Holds(control,W,T-1)]$$

We are now ready to infer *B*'s duty to stop at time *1* i.e. *Must(stop, B, 1) (*with *V = A, W = B, T = 1) :*

$Holds(combine(follows,V),W,T) \wedge Holds(stop,V,T) \rightarrow Must(stop,W,T)$. The meaning of this rule is: if *W* follows *V* in a file at time *T*, and at that time *V* stops, then *W* must stop too in order to avoid a crash.

To infer the ability of *B* to stop at *T*, we use the following basic inference rule:

$$Able(E,A,T) \leftrightarrow (\exists Act) Action(Act) \wedge Pcb(Act,E) \wedge Available(Act,E,A,T)$$

This rule means that an agent *A* is able to reach some effect *E* at time *T*, if and only if there is some action *Act* that is a potential cause of *E* (*Pcb* means "potentially caused by"), and *Act* is available to *A* to reach *E* at time *T*. The set of actions, effects and potential causes are stored in static data bases (for example, the data base contains *Pcb(brake, stop)* to express that stopping is potentially caused by braking). Moreover, we have a default which states that in general, actions are available for the agents to reach the corresponding effects. This rule has a number of exceptions expressed by material implications that inhibit the default [2]. In our case, none of the exceptions is verified. Thus, we obtain: *Available (brake, stop, B, 1)* and consequently *Able(stop, B, 1)*.

Finally, by applying the first form of a primary anomaly, we can infer the predicate *P_Anomaly* and the cause of the accident is that "*B* did not stop at a time where s/he had to stop".

## VI. Implementation

To implement the reasoning system, we are using *SMODELS*[2], an answer set programming language based on the stable model semantics [1]. To give a general idea about the method used to transform default logic rules into *SMODELS* rules we consider the following simple cases where *A, B, C* are reified first order literals[3].

- A material implication $A \rightarrow B$ is translated into the couple of rules: *B :- A.* and *–A :- –B* (for contraposition)
- A normal (resp. semi-normal) default *A : B* (resp. *A : B [C]*) is transformed into the rule: *B :- A, not –B.* (resp. *B :- A, not –B, not –C.*)

We have tested our approach on a corpus of 60 short texts (the average length of the texts of the corpus is about 3 lines). For each text, the reasoning system gives successfully the desired primary and derived anomalies. The number of inference rules used actually in the reasoning system is about 200 rules and the answer time varies according to the text between 6 and 30 seconds. Among other things, the answer time depends on the number of time intervals and the number of agents considered in a given text. The former number varies in the corpus between 2 and 6 time intervals whereas the second one varies between 1 and 4 agents.

## VII. Conclusion and perspectives

We propose in this work a non-monotonic reasoning system that uses the norms of the car-driving domain to infer automatically the cause of an accident from its textual description. The relationship between the notions of norm and cause is established by considering the cause of the accident as being the most specific norm which has been violated in the text. The next step of our work is to complete the validation of the approach on the remainder of the corpus; then we will finish the implementation of the last part of the system which deals with the linguistic reasoning. We hope in a longer term perspective to generalize the approach to other domains and to explore the idea of indexing textual documents using the norms of their domains.

## References

[1] Gelfond, M and Lifschitz, V. The Stable Model Semantics for logic programming. In *Proceedings/Actes, 5th International Conference on Logic Programming*, pages 1070-1080, 2004.

[2] Kayser, D and Nouioua, F. Representing Knowledge about Norms, In *Proceedings/Actes, 16th ECAI Conference*, pages 363-367, 2004.

[3] Kayser, D and Nouioua, F. About Norms and Causes, *International Journal on Artificial Intelligence Tools*. Special Issue on FLAIRS 2004, 14(1 & 2), To appear, 2005.

[4] McDermott, D.V. A Temporal Logic for Reasoning about Processes and Plans *Cognitive Science* 6, pages 101-155, 1982.

[5] Reiter, R. A Logic for Default Reasoning, *Artificial Intelligence, Special Issue on Nonmonotonic Logic*, 13(1-2), pages 81-132, 1980.

[6] Schank, R.C and Abelson, R.P. *Scripts, Plans, Goals and Understanding* Lawrence Erlbaum Ass, 1977.

---

[2] *SMODELS* and its front-end *LPARSE* are available in the web page: http://www.tcs.hut.fi/Software/smodels/
[3] '–' stands for the hard negation and '*not*' stands for negation by failure.